\documentclass{article} % For LaTeX2e
\usepackage[preprint]{colm2026_conference}
\usepackage{listings}
\usepackage{microtype}
\usepackage{hyperref}
\usepackage{url}
\usepackage{booktabs}
\usepackage{graphicx}
\usepackage{tabularx}
\usepackage{subfigure}
\usepackage{algorithm}
\usepackage{multirow}
\usepackage{amsmath}
\usepackage{tcolorbox}
\usepackage{colortbl}
\usepackage{algpseudocode}
\usepackage{latexsym}
\definecolor{darkgreen}{RGB}{50,100,0}
\usepackage{pifont}
\usepackage{amssymb}
\usepackage{wrapfig}
\usepackage{enumitem}
\usepackage{lineno}
\newcommand\logo{\raisebox{-5pt}{\includegraphics[width=7.5em]{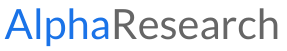}}}
\definecolor{darkblue}{rgb}{0, 0, 0.5}
\hypersetup{colorlinks=true, citecolor=darkblue, linkcolor=darkblue, urlcolor=darkblue}
\usepackage{xspace}
\usepackage[textsize=tiny]{todonotes}
\newcommand{\ours}{AlphaResearch\xspace}
\newcommand{\dataset}{AlphaResearchComp\xspace}
\title{\logo{}: Can Peer-reviewed Language Models Accelerate New Algorithm Discovery?}

\author{Zhaojian Yu$^{1}$\thanks{~~Equal contribution.},  
Kaiyue Feng$^{2*}$,  Yilun Zhao$^3$, Shilin He,  Xiao-Ping Zhang$^{1}$\thanks{~~Corresponding author.},  Arman Cohan$^3$
\\
$^1$Tsinghua University
$^2$New York University
$^3$Yale University
\\
[0.3em]
\makebox[\linewidth]{\href{https://github.com/answers111/alpha-research}{https://github.com/answers111/alpha-research}}
}
\begin{document}

\ifcolmsubmission
\linenumbers
\fi

\maketitle
\vspace{-6pt}
\begin{abstract}
LLMs have made significant progress in complex but easy-to-verify problems, yet they still struggle with discovering the unknown.
In this paper, we present \textbf{AlphaResearch}, an autonomous research agent designed to discover new algorithms on open-ended problems by iteratively running the following steps: (1) propose new ideas (2) program to verify  (3) optimize the research proposals.
To synergize the feasibility and innovation of the discovery process, we construct a novel dual environment by combining the execution-based verifiable reward and reward from simulated real-world peer review environment in AlphaResearch.
We construct \textbf{\dataset}, a set of questions that includes an eight open-ended algorithmic problems competition to benchmark AlphaResearch.
Experimental results show that AlphaResearch achieves stronger discovery performance than other agentic discovery systems on six open-ended problems.
Notably, the algorithm discovered by AlphaResearch on the \emph{``packing circles''} problem achieves the best-of-known performance, surpassing the results of human researchers and strong baselines from recent work (e.g., AlphaEvolve). 
Additionally, we conduct a comprehensive analysis of the benefits and remaining challenges of autonomous research agent, providing valuable insights for future research.
\end{abstract}

\begin{figure*}[h]
    \centering
    \includegraphics[width=0.95\linewidth]{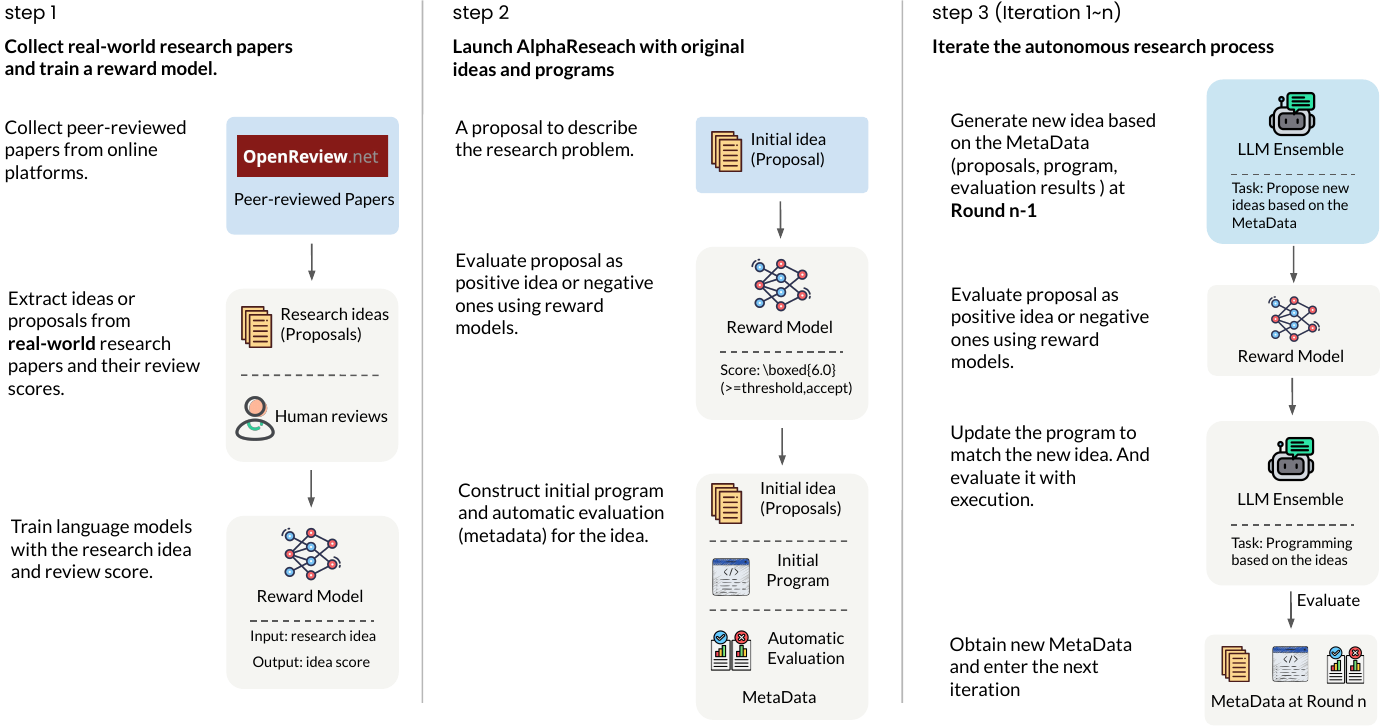} 
    \caption{
    % The launch of AlphaResearch contains two steps: (1) Train reward models with real-world peer-reviewed records. (2) Prepare initial research proposals, initial programs and evalution program. It will then iteratively refine the research proposals and programs (3).
    Overview of AlphaResearch. This system accelerates discovery process with interleaved idea generation and program generation, where it get reward from peer-review model and program execution.
    }
    \label{fig:overview}

\end{figure*}

\section{Introduction}
Recent progress has shown that sophisticated coding agent scaffolds \citep{yang2024swe} could help frontier LLMs~\citep{gpt5, comanici2025gemini} achieve expert-level performance in end-to-end coding problems (e.g., modify the code to pass the test cases in sandbox~\citep{jimenez2024swebench}).
% But for open-ended coding problems where the final results (end-points) are unpredictable, end-to-end coding agents often 
However, for open-ended coding problems where the target endpoint (i.e., the optimal implementation) is left underspecified, end-to-end coding agents often struggle to converge on the best possible solution with just a few attempts.

AlphaEvolve~\citep{novikov2025alphaevolve} shows that scaling the number of agent attempts can reveal an emerging ability to derive stronger solutions from suboptimal prior results. 
% However, it relies on a "brute-force search" system: each new attempt is generated by sampling a previous attempt from the program database, and this process repeats until a preset iteration limit is reached.
% This leads to a significant increase in runtime for validating attempts, which naturally raises the following research question:
However, its brute-force search process repeatedly samples new attempts from a program database until a preset iteration limit is reached, substantially increasing validation cost.
This raises a key question:
\emph{Can we accelerate new algorithm discovery by reducing ineffective attempts?}

ShinkaEvolve~\citep{lange2025shinkaevolve} advances this direction by introducing a more sample-efficient evolutionary framework for open-ended program evolution. Specifically, it filters out redundant or minimally changed edits before expensive validation using a novelty rejection-sampling mechanism that combines embedding-based code similarity with an LLM-based novelty judge. This improves exploration efficiency relative to pure brute-force sampling, but important limitations remain. Embedding-based similarity thresholds can miss semantically meaningful changes or, conversely, fail to detect subtle redundancies. In addition, LLM-based novelty judgments are inherently unreliable because there is little high-quality training data explicitly annotated for program novelty in open-ended evolutionary search. As a result, general-purpose LLMs may rely on superficial cues and produce judgments that diverge from expert assessments of functional improvement, algorithmic originality, or non-trivial semantic change~\citep{zheng2023judging, szymanski2025limitations, lin2025evaluating}. These limitations motivate the need for more robust and efficient mechanisms for balancing exploration and exploitation in scalable algorithm discovery systems.
In this paper, we introduce \textbf{\ours}, an autonomous research-based coding agent that could reduce ineffective attempts and accelerate new algorithm discovery by interacting with real-world peer-review  environments.
As shown in \autoref{fig:overview} and \autoref{tab:compofagent},  AlphaResearch construct a novel dual research-based environment, where the research ideas proposed by LLMs could be judged by a simulated real-world peer-review environment and then attempt by code sandbox.
Specifically, we (1) train a reward model \textbf{AlphaResearch-RM-7B} with real-world peer-review records, addressing the limitation of prior coding-only approaches that lack real-world research feedback, and use it to score the fresh ideas generated by LLMs;
\begin{table}[!t]
    \centering
    \small
    \begin{tabular}{l|ccc}
    \toprule
\multirow{2}{*}{\textbf{Agent}} 
    & \multicolumn{3}{c}{\textbf{Environments}} \\
    & Sandbox                  & Database & Peer-review Feedback \\
    \midrule
        SWE-agent \citep{yang2024swe} & \color{darkgreen}\ding{52} & \color{red}\ding{56} & \color{red}\ding{56} \\
        AlphaEvolve \citep{novikov2025alphaevolve} & \color{darkgreen}\ding{52} & \color{darkgreen}\ding{52} & \color{red}\ding{56} \\
        ShinkaEvolve \citep{lange2025shinkaevolve} & \color{darkgreen}\ding{52} & \color{darkgreen}\ding{52} & \color{red}\ding{56} \\
        AlphaResearch (ours) & \color{darkgreen}\ding{52} & \color{darkgreen}\ding{52} & \color{darkgreen}\ding{52} \\
      \bottomrule
    \end{tabular}
    \caption{Comparison of AlphaResearch and other agents.}
    \label{tab:compofagent}
\end{table}
% Specifically, we (1) train a reward model \textbf{AlphaResearch-RM-7B} with real-world peer-reviewed records to simulate the real-world peer review environment and score the fresh ideas generated by LLMs 
% \ac{This is a good contribution, but we don't motivate it well (e.g., what gap exists in current approaches) in the previous paragraph. } 
(2) construct an automatic program-based verifiable environment that executes these ideas with an interpreter.
This dual environment facilitates a rigorous algorithm discovery process for autonomous research agents. As illustrated in \autoref{fig:overview}, AlphaResearch discovers new algorithms by iteratively running the following steps: (i) proposing new research ideas, (ii) verify the ideas in the dual research-based environment, and (iii) optimizing the proposals for higher reward from the environment. 
The synergy between an iterative real-world peer review environment and program-based verification empowers AlphaResearch to continuously explore novel research ideas and verify the attempts via less program execution. 

% \ac{we need to motivate why we create this and why existing resources and datasets (e.g., those used in AlphaEvolve) are not sufficient to evaluate AlphaResearch}
To facilitate the comparison between AlphaResearch and other baselines (e.g., AlphaEvolve and ShinkaEvolve), we collect 8 open-ended research problems and their best-of-human records \textbf{\dataset} and simulate an algorithm discovery \textbf{comp}etition between AlphaResearch and other agents for discovery.
% \ac{Why we name it AlphaResearchComp? (Comp part is unclear).}
% \ac{Need to add one more sentence to describe what is special about this dataset, what are the composition of problems, etc.}
Our experimental results show that AlphaResearch achieves a score of 2.939 on the \emph{“Packing Circles (n=32)”} problem, surpassing the previous results obtained by AlphaEvolve (see \autoref{app:examples}). 
Furthermore, compared with ShinkaEvolve, which adopts code-novelty-based rejection sampling, AlphaResearch demonstrates faster convergence to higher-performing programs on three out of four open-ended problems. This highlights the effectiveness of the peer-review environment constructed in AlphaResearch. 
We also analyze the advantages and remaining challenges of autonomous research agents for knowledge discovery, providing insights for future work.
% Our results demonstrate that AlphaResearch surpasses human researchers on two problems but fails on the other six. 

% In the competition with OpenEvolve~\citep{openevolve}, an open-sourced implementation of AlphaEvolve,
% AlphaResearch optimizes the result of \emph{``Packing Circles (n=32)''} problem to 2.939, where the goal is to pack $n$ disjoint circles inside
% a unit square so as to maximize the sum of their radii, surpassing the results of best-of-human and previous SoTA results achieved by AlphaEvolve (as shown in \autoref{app:examples}).
% Furthermore, compared to AlphaEvolve, AlphaResearch's discovery process exhibits better and faster convergence on other six competitions. 
% \ac{These results are not very meaningful if people don't understand the metric and the problem. Maybe don't go into details of the numbers here.}
% \ac{we need to also add description saying that while it works on 2 problems, it still fails on 6 of the other problems, showing that there is large room for future work. And then say that we perform careful analysis to provide useful insights...}

To summarize, our key contributions are the following: 
\begin{itemize}[leftmargin=*]
\itemsep0em
    \item We introduce AlphaResearch, a novel autonomous research agent designed to accelerate the discovery of new algorithms by synergistically combining idea generation, code implementation, and verification in a dual environment of simulated peer review and executable program validation.
    \item We benchmark AlphaResearch with other agents for discovery across diverse verifiable problems to showcase the advantages of AlphaResearch in reducing ineffective attempts and accelerating new algorithm discovery.
    \item We present systematic ablations and analysis to understand the importance of incorporating peer-review environments into LLM-based autonomous discovery system.
\end{itemize}

% \newpage

\section{Related Work}
% \paragraph{Accelerating Scientific Discovery with LLMs.}
% Recent work demonstrates that LLMs have the potential to break the scientific boundary built by human researchers. For example, AlphaEvolve \citep{novikov2025alphaevolve}, as an evolutionary coding agent, developed a search algorithm that found a procedure to multiply two 4×4 complex-valued matrices using 48 scalar multiplications, improves Strassen (1969)’s algorithm.
% ASI-ARCH \citep{liu2025alphago} discovers 106 state-of-the-art linear attention architectures by combining LLMs with neural architecture search (NAS). AlphaResearch also finds the best-of-known \emph{``Packing Circles''} solutions, improving the previous algorithm discovered by human and AlphaEvolve.
\paragraph{LLMs for New Ideas.}
Several recent works explored methods to improve research idea generation, such as iterative novelty refinement \citep{wang2024scimon, Baek2024ResearchAgentIR}. These works focus on improving the research idea over vanilla prompting but critically miss an effective verification method. To promote more reliable AI-generated research ideas, many studies have proposed solutions from different perspectives, such as comparisons with any human expert \citep{si2024can}, using LLMs for executing experiments by generating code with human-curated research problems \citep{Huang2023MLAgentBenchEL, Tian2024SciCodeAR}, and executing LLM-generated research ideas with LLM-generated programs \citep{Li2024MLRCopilotAM, lu2024ai, aygun2025ai}. These works either use automatic program evaluation or unverifiable LLM evaluator method, which presents a challenge for their scalability to real-world advanced algorithm discovery. 
Our AlphaResearch presents a more feasible direction by combining program execution with RM training from real-world peer-reviewed research records.

\paragraph{LLMs for Code Generation.}
In autonomous research agents, code generation serves as a fundamental step. Previous models~\citep{guo2024deepseek, yu2023wavecoder, hui2024qwen2} and benchmarks ~\citep{chen2021evaluating,yu-etal-2025-humaneval} for code generation are in a longstanding pursuit of synthesizing code from natural language descriptions. SWE-Bench \citep{jimenez2024swebench}, PaperBench~\cite{starace2025paperbench}, MLE-Bench~\cite{chan2024mle} introduces the problems in real-world agentic coding. Many studies on SWE-Bench have greatly contributed to the emergence of coding agents like SWE-Agent \citep{yang2024swe} and OpenHands \citep{wang2025openhands}. These agent frameworks greatly facilitate the training of agentic LLMs like Kimi-K2 \citep{team2025kimi} and GLM-4.5 \citep{zeng2025glm}.
The surge of these models on SWE-Bench underscores a critical need to reassess the future directions of coding agent research.
% Our AlphaResearchComp benchmark shows that testing LLMs on open-ended research for algorithm discovery is a promising direction to adapt language models to real-world tasks.

\section{AlphaResearch}

\subsection{Overview}

% \ac{It is important to clearly mention what is novel about AlphaResearch in this paragraph and position it against existing work.}
AlphaResearch accelerates algorithm discovery process by continuously optimizing the research outcome from the dual reward that synergizes rigorous program verification and a real-world peer review environment. 
As shown in \autoref{fig:overview}, given initial idea $i_0$ and program $p_0$, AlphaResearch runs the program $p_0$ with execution, producing $r_0$, which represents the initial overall rating. The triplet $(i_0, p_0, r_0)$ will be fed to AlphaResearch for subsequent processing, including newer idea generation, code implementation, and program-based execution. When reaching a point where execution output $r_n$ surpasses the previous rating, AlphaResearch will save the triplet $(i_{best}, p_{best}, r_{best})$ as the best record. We repeat the process until $r_{best}$ surpasses the best-of-human score, or the maximum round is reached. The resulting trajectory is denoted as $\tau = i_0p_0r_0...i_{n-1}p_{n-1}r_{n-1}i_{n}p_{n}r_{n}$, where $n$ is the total rounds.

\subsection{Actions}    
\paragraph{New Idea Generation.} For each step $k$, AlphaResearch start with generating a new idea $i_k$ based on a sampled previous step  $(i_{t}, p_{t}, r_{t})$ from previous trajectory $\tau_{k-1}=i_0p_0r_0...i_{k-1}p_{k-1}r_{k-1}$. This process can be denoted as:
\begin{equation}
i_k\thicksim\mathbb{P}_\mathcal{A}(\cdot|i_{t}\oplus p_{t}\oplus r_{t})
\end{equation}
where $\oplus$ means concatenation, $t$ is the sampled step from trajectory $\tau_{i-1}$ and $\mathbb{P}_\mathcal{A}(·)$ indicates random sampling. We use a reward model to select high-quality ideas overall. If $\mathcal{RM}(i_n)$ outputs a negative score, we cease the subsequent actions in this round.

\paragraph{Program-based Verification.}
After obtain the fresh idea, AlphaResearch generates new program $p_k$ based on the previous implementation $p_t$ and new idea $i_k$ next:
\begin{equation}
p_k\thicksim\mathbb{P}_{\mathcal{A}}(\cdot|p_t \oplus i_k)
\end{equation}
and yield the evaluation result $r_k$ by verifying $p_k$ with code executor $r_k\leftarrow\mathcal{E}(p_k)$. Then, we update the trajectory $\tau_k$ with the newly generated idea $i_k$, program $p_k$ and result $r_k$:
\begin{equation}
\tau_k\leftarrow\tau_{k-1}\oplus i_k\oplus p_k\oplus r_k
\end{equation}
We repeat the above interaction process until $k$ reaches the maximum rounds $n$ and get the best result $(i_{best}, p_{best}, r_{best})$ as final output.

\begin{algorithm}[t]
\small
\caption{AlphaResearch}
\label{alg:inference}
\textbf{Require:} initial idea $i_0$, initial program $p_0$, initial result $r_0$, model $\mathcal{A}$, evaluation program $\mathcal{E}(\cdot)$, maximum iteration rounds $n$, 
\begin{algorithmic}[1]
\State $\tau_0 \leftarrow (i_0, p_0, r_0)$, $r_{best}=0$ \Comment{Initialization}
\For{$k=1$ to $n$ do}
\State $(i_t, p_t, r_t) \sim \mathbb{P}(\cdot|\tau_{k-1})$ \Comment{States Sampling}
\State $i_k\thicksim\mathbb{P}_\mathcal{A}(\cdot|i_{t}\oplus p_{t}\oplus r_{t})$ \Comment{New Idea Generation (Eq. 1)}
\If{$\mathcal{RM}(i_k)$ < threshold} 
\State \textbf{continue} \Comment{Reward Model for New Idea}
\EndIf
\State $p_k\thicksim\mathbb{P}_{\mathcal{A}}(\cdot|p_t \oplus i_k)$ \Comment{Program Generation (Eq. 2)}
\State $r_k\leftarrow\mathcal{E}(p_k)$ \Comment{Program-based Execution}
\If{$r_k$ $>$ $r_{best}$}
\State $(i_{best}, p_{best}, r_{best})$ = $(i_{k}, p_{k}, r_{k})$
\EndIf
\State $\tau_k\leftarrow\tau_{k-1}\oplus i_k\oplus p_k\oplus r_k$ \Comment{Trajectory Update (Eq. 3)}
\EndFor
\State \textbf{return} $(i_{best}, p_{best}, r_{best})$
\end{algorithmic}
\end{algorithm}

\subsection{Environment}

\subsubsection{Reward from Real-world Research Records}
\label{sec:rm}
Existing autonomous idea generation process suffers from a trade-off where highly novel research ideas may lack feasibility \citep{guo2025ideabench, si2025ideation}. To address this gap and ensure the feasibility of idea candidates, we train a reward model with ideas from real-world peer-review information to simulate the real-world peer-review environment.

\begin{wraptable}{r}{0.47\columnwidth}
\centering
\vspace{1.0\baselineskip}
\small
\setlength{\tabcolsep}{4pt}
\begin{tabular}{l|c|c}
\toprule
\textbf{Split} & \textbf{Train} & \textbf{Test} \\
\midrule
\textbf{Data Source} & ICLR & ICLR \\
\textbf{Range of Date} & 2017$\sim$2024 & 2025 \\
\textbf{Environment Nums} & 24,445 & 100 \\
\textbf{Start Date} & 2016-11 & 2024-10 \\
\textbf{End Date} & 2023-12 & 2024-12 \\
\bottomrule
\end{tabular}
\caption{Dataset for reward model training. We use the end of author-reviewer rebuttal period as the latest knowledge date.}
\label{tab:rmd}
\end{wraptable}

\paragraph{Dataset for reward model.} 
To train RM to identify good ideas, we collect all ICLR peer review records from 2017 to 2024 as our training set. We sample a subset of ICLR 2025 records as a test set, where the dates of train and test are disjoint, which prevents knowledge contamination between the train and test split. We also select Qwen2.5-7B-Instruct\footnote{Its release date \texttt{2024-09} is earlier than the ICLR 2025 author-reviewer rebuttal period \texttt{2024-10}.} as our base model.
For each record, we extract the abstract part as RM input and wrap the average peer-review overall ratings with \texttt{\textbackslash boxed\{\}} as RM output. We fine-tune the model with RM pairs, yielding the AlphaResearch-RM-7B model.

\paragraph{RM threshold.}
To simplify the RM evaluation, we binarize the RM output score according to the ICLR Reviewer Guide, where overall rating $> 5.5$ records are regarded as a positive score and $\leq 5.5$ records are negative. We do not treat the RM threshold as a hyperparameter in this work, as the score carries real-world interpretability where it corresponds to the midpoint between the acceptance and rejection scores at ICLR.
We compute the binary classification accuracy and evaluate three models (GPT-5, Qwen2.5-Coder-Instruct, and AlphaResearch-RM-7B) on the AlphaResearch-RM test set.

\begin{wraptable}{r}{0.47\columnwidth}
  \centering
 % \yilun{also report human annotator performance? (might be close to our model)}
   \label{tab:rm-res}
  \begin{tabular}{l|cl}
  \toprule
  \textbf{Reward Model} & \textbf{Cutoff} & \textbf{Acc} \\
  \midrule
  Random (theoretical) & - & 50.0\% \\
  Human Annotator & - & 65.0\% \\
  \midrule
  % Deepseek-V3-0324 & 2025-03 & 39.0\% \\
    GPT-5 (medium) & 2025-08 & 53.0\% \\
  Qwen2.5-7B-Instruct & 2024-09 & 37.0\% \\
  AlphaResearch-RM-7B & 2024-09 & 72.0\% \\
  \bottomrule
  \end{tabular}
\caption{Evaluation results of different RMs. We use the more recent date between the model release date and the dataset cutoff as the latest date.}
\end{wraptable}

\paragraph{Can LLMs identify good ideas?} To establish a human annotator baseline, we select 3 researchers with relevant backgrounds who have published papers and served as a reviewer on their assigned topics.
\autoref{tab:rm-res} presents the evaluation results that eliminate the knowledge contamination, highlighting the following observations: (1) Both GPT-5 and Qwen2.5-7B-Instruct achieve lower than 60\% accuracy when identifying the good ideas from ICLR 2025 records.
(2) After being fine-tuned with ideas from previous ICLR peer-review information, AlphaResearch-RM-7B demonstrates 72\%
% (>50\%, above the accuracy of random prediction) 
binary classification accuracy on unseen ICLR 2025 ideas, significantly outperforming baseline models and human annotators. 
Based on these observations, we use the fine-tuned AlphaResearch-RM-7B as the final RM to simulate a real-world peer-review environment and filter out good ideas generated by AlphaResearch.

\subsubsection{Reward from Program-based Execution}
We construct an automatic evaluation process with a code executor where each new program $p_k$ generated by AlphaResearch will be captured and evaluated. 
The evaluation program $\mathcal{E}(\cdot)$ includes two modules: (i) \textbf{Verification} module that validates whether $p_k$ conforms to the problem constraints. (ii) \textbf{Measurement} module that output the score $r_k$ of program performance.
The program output $r_k$ will be injected into the idea generation prompt (if sampled), thereby participating in the optimization process for fresh ideas. These programs and results are stored in a candidate pool, where the primary goal is to optimally resurface previously explored ideas in future generations. The verifiable reward by code executor significantly simplifies the action spaces of AlphaResearch, thereby enhancing the efficiency of the discovery process.

\section{Experiments}

\subsection{Setup}
We select \texttt{o4-mini}, a strong but cost-efficient LLM as our research agent and run AlphaResearch on each problem to get the best algorithm. We perform supervised finetuning on \texttt{Qwen-2.5-7B-Instruct} \citep{yang2025qwen3} with the collected ICLR records, yielding AlphaResearch-RM-7B. We do not compute loss on paper information, only on the average rating scores within \texttt{\textbackslash boxed\{\}}. For fine-tuning hyperparameters, we train our model with a learning rate of 1e-5 warmed up linearly for 100 steps. We train all the models in bfloat16 precision with Pytorch
Fully Shard Data Parallel (FSDP) and set a global batch size to 128 for 2 epochs.
All other settings not mentioned in this paper follow the default values
of Huggingface Trainer \footnote{\url{https://huggingface.co/docs/transformers/main_classes/trainer}}. Due to the unavailability of the AlphaEvolve codebase, we adopted OpenEvolve and ShinkaEvolve as our baseline approaches.

\subsection{Evaluation}
\begin{table}[t]
% \resizebox{\linewidth}{!}{%
\centering
\resizebox{\linewidth}{!}{%
\begin{tabular}{l|cc|cc|c}
\toprule
\multirow{2}{*}{\textbf{Problem}}                        & \multicolumn{2}{c|}{\textbf{Human Researcher}} & \multicolumn{2}{c|}{\textbf{AlphaResearch}} & \multirow{2}{*}{\textbf{$\Delta$(\%)}} \\
                                                & \textit{research record}  &  \textit{baseline}   & \textit{init}            & \textit{best}            &            \\
\midrule
packing circles (n=26) $\uparrow$   &  D. Cantrell (2011)  & 2.634    &  0   & 2.636  & 0.002   \\
packing circles (n=32) $\uparrow$   & E. Specht (2012)    & 2.936     &  0  & 2.939  &  0.003  \\
minimizing max-min distance ratio $\downarrow$ & D. Cantrell (2009) & 12.89 & 15.55 & 12.92 & - 0.03\\
third autocorrelation inequality   $\downarrow$  & C. Vinuesa  (2009)  & 1.458  & 35.746 & 1.546 & -0.088   \\
% kissing number                                  & 592           &                  \\
spherical code  (d=3, n=30)      $\uparrow$      &   Hardin \& Sloane (2002)     &    0.6736  & 0.5130 & 0.6735 & -0.0001\\
autoconvolution peak minimization $\downarrow$  &  Matolcsi \& Vinuesa (2010)  & 0.755   & 1.512 &      0.756     & -0.001   \\
littlewood polynomials (n=512) $\downarrow$  & Rudin \& Shapiro (1959)  & 32     &32& 32  &  0 \\
MSTD        (n=30) $\uparrow$ & Hegarty (2007)    & 1.04   &1.04    &  1.04 & 0   \\

\bottomrule
\end{tabular}
}
\caption{Results on AlphaResearchComp. $\uparrow$ indicates that higher score is better and $\downarrow$  for lower. $\Delta$ indicates the performance gap between best of AlphaResearch and human baseline.}
\label{tab:res}
\end{table}

\begin{figure}[t]
    \centering
    \includegraphics[width=0.40\linewidth]{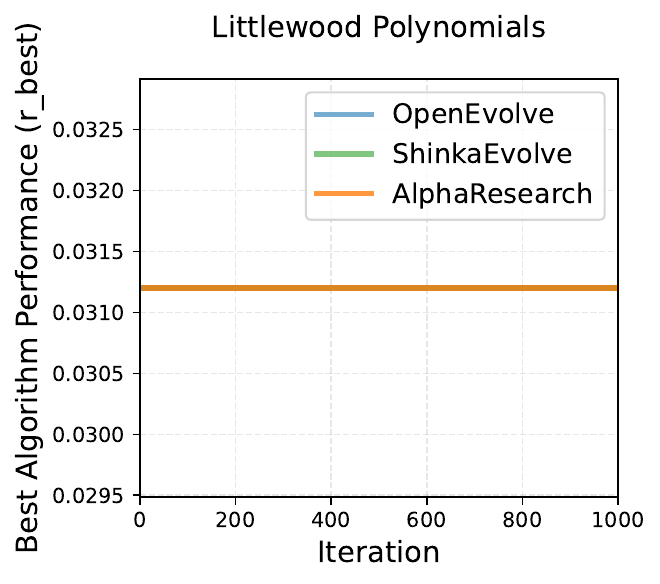}
    \includegraphics[width=0.40\linewidth]{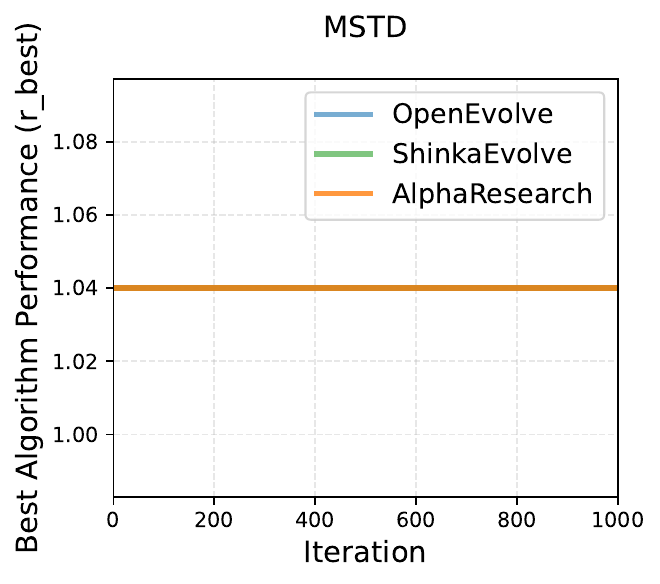}
    \caption{Performance comparison of AlphaResearch, OpenEvolve, and ShinkaEvolve in terms of failure modes of AlphaResearchComp.}
    \label{app:comp2}
\end{figure}

\paragraph{Problem collection.}
We curate AlphaResearchComp,  a set of frontier program-based research tasks including geometry, number theory, harmonic analysis, and combinatorial optimization. These problems were selected based on the following principles: (1)  Each task has a precise mathematical formulation with an objective function that admits rigorous automatic evaluation. 
(2) For every problem, we provide the best-known human result from the literature. These represent conjectured best-known values rather than proven optima, ensuring ample room for further improvement. 
The curated problems are either inherited from prior work (e.g., AlphaEvolve) or collected from online repositories and domain experts. Each problem is supported by verifiable resources in the corresponding field. 
This design enables AlphaResearch to demonstrate both the \emph{reproducibility} of established mathematical results and the \emph{potential for discovery} beyond current human-best achievements.

\subsection{Main Results}
\paragraph{Successful Cases.}

% \textcolor{blue}{\textbf{LLMs could sometimes discover new algorithms themselves.} }
% \textbf{The discovery of superhuman algorithms remains challenging for LLMs.}
\autoref{tab:res}
% \kaiyue{probably want to check the autorefs in this section, I guess you mean Table 4 here?}
presents the results of AlphaResearchComp on 8 algorithms discovery problems. AlphaResearch achieves a 2/8 win rate ($\Delta$ $>$ 0) against human researchers, with one notable success: the algorithm discovered by AlphaResearch for \emph{``Packing Circles''} problem reaches the best-of-known performance (2.636 for n=26, 2.939 for n=32), outperforming human researchers (2.634 for n=26, 2.936 for n=32) and AlphaEvolve (2.635 for n=26, 2.937 for n=32), where the case (n = 32) is shown in \autoref{fig:comp-alpha}.

\paragraph{Failure Cases.}

% Despite exhibiting continuous growth, AlphaResearch’s performance still underperforms human researchers in \emph{``Littlewood polynomials``} and \emph{``MSTD(n=30)``} problems, where AlphaResearch have not shown an increase in execution-based rewards. This indicates that current LLMs still struggle to consistently find better algorithms than human researchers.

Although AlphaResearch has exhibited continuous growth, it still underperforms human researchers on the \emph{``Littlewood polynomials``} and \emph{``MSTD(n=30)``} tasks, with no observable improvement in program score throughout the discovery process. 
To further analyze this limitation, we compare the performance of AlphaResearch with that of OpenEvolve and ShinkaEvolve on the failure modes.
As shown in \autoref{app:comp2}, All of AlphaResearch, OpenEvolve, and ShinkaEvolve fail to improve on the Littlewood polynomials and MSTD (n=30) problems, which indicates that current large language models still face significant challenges in reliably discovering superior algorithms.

\begin{figure}[t]
    \centering
    \includegraphics[width=0.42\linewidth]{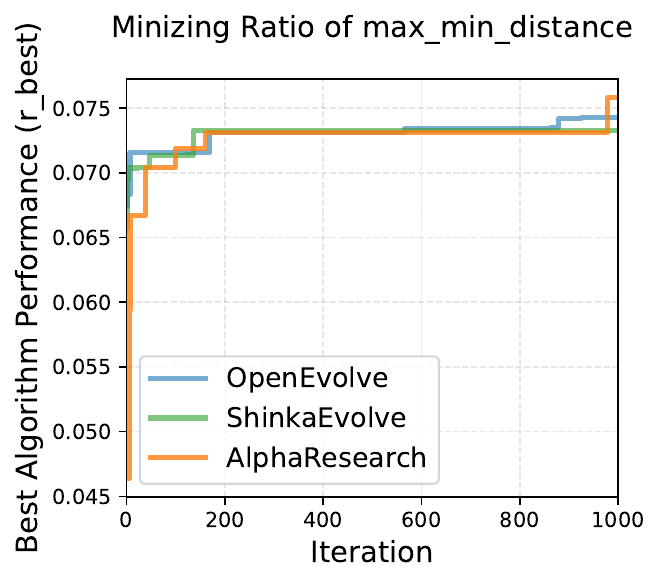}
    \includegraphics[width=0.42\linewidth]{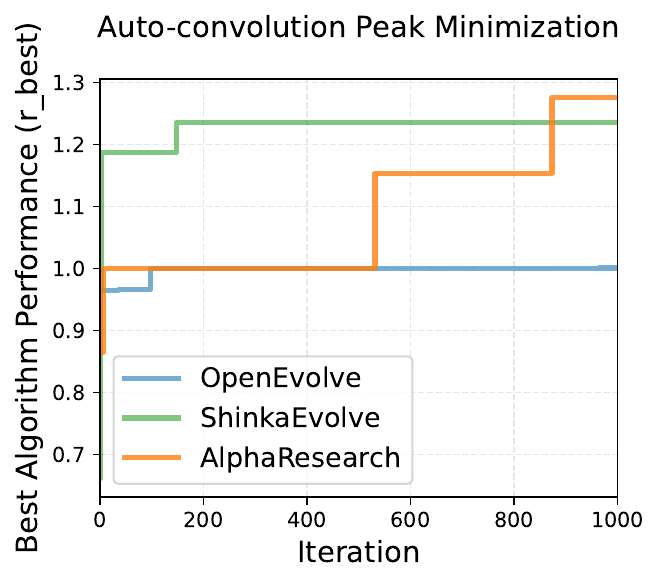}
    \includegraphics[width=0.42\linewidth]{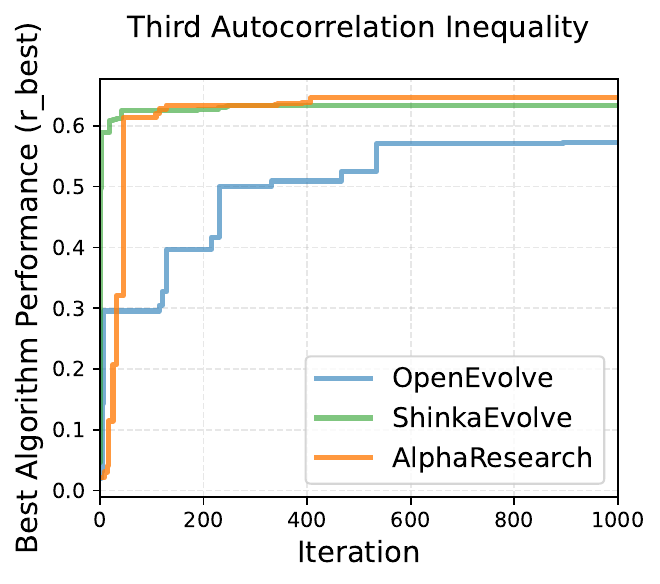}
    \includegraphics[width=0.42\linewidth]{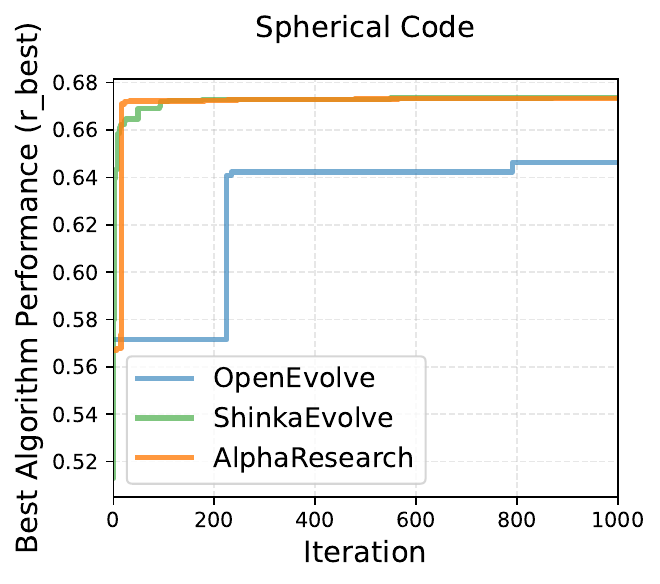}
    \caption{Comparison of AlphaResearch, OpenEvolve, and ShinkaEvolve throughput the 1000 step discovery process. }
    \label{app:comp}
\end{figure}
\subsection{Comparison with OpenEvolve and ShinkaEvolve}

\autoref{app:comp} presents the comparison of AlphaResearch with OpenEvolve and ShinkaEvolve, highlghting the following observations:
(1) With interleaved idea generation and program generation, AlphaResearch consistently surpasses OpenEvolve across the first four open-ended problems, which demonstrates the effectiveness of incorporating peer-review environments into autonomous discovery process.
(2) Compared to ShinkaEvolve, which employs code novelty-based rejection sampling to reduce the number of attempts, AlphaResearch surpasses it on three out of four tasks within the first 1,000 iterations. Although the initial performance gains are modest, these advantages tend to accumulate into more substantial improvements as the number of iterations increases. This highlights the superiority of the peer-review environment over approaches that rely solely on code novelty-based rejection sampling.

\subsection{Ablations and Analysis}

% \begin{wrapfigure}{r}{0.47\linewidth}
%     \centering
%     \includegraphics[width=\linewidth]{figures/dis.pdf}
%     \caption{The idea comparison between the execution-only research agent and AlphaResearch, where AlphaResearch-RM-7B is used. \textcolor{blue}{This is done between the full distribution of all 1000 generated ideas from both agents without filtering.}}
%     \label{fig:dis}
%     \vspace{-8pt}
% \end{wrapfigure}

\begin{wrapfigure}{r}[0pt]{0.4\linewidth}
    \centering
    \includegraphics[width=\linewidth]{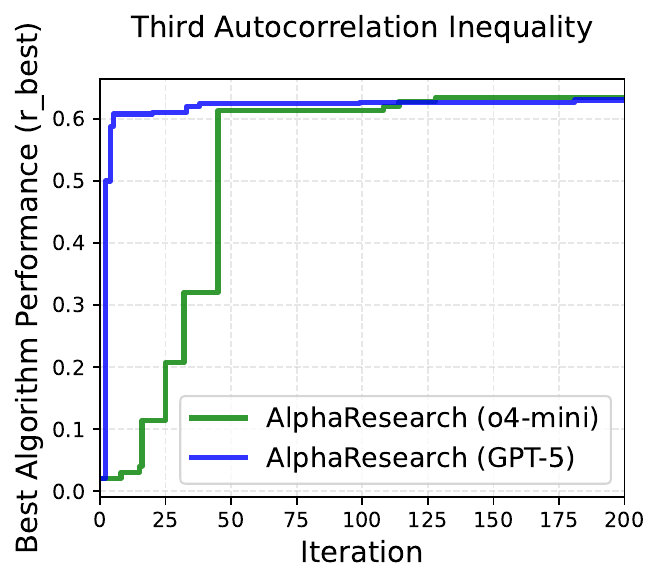}
    \caption{Comparison between different frontier LLMs in AlphaResearch.}
    \label{fig:gpt5}
    \vspace{-6pt} 
\end{wrapfigure}

\textbf{Impact of different LLM backbone.}
To compare the impact of different LLM backbones on AlphaResearch, we used GPT-5 and o4-mini to run AlphaResearch for 200 steps on the "The Autocorrelation Inequality" problem, respectively.
As illustrated in \autoref{fig:gpt5}, AlphaResearch (GPT-5) achieves strong performance much more rapidly than o4-mini during the early stages of discovery. However, in later stages, the two models exhibit comparable performance, suggesting that their underlying capabilities on the algorithm discovery task are similar.
\begin{figure*}[t]
    \centering
    \includegraphics[width=\linewidth]{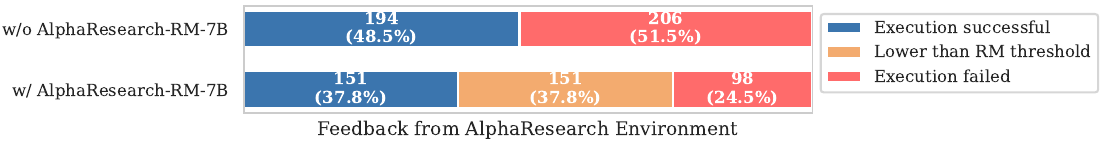}
    \caption{The impact of real-world peer review environment on execution results. AlphaResearch-RM-7B filters 151 bad ideas, where 108 ideas fail to execute and 43 are successful.}
    \label{fig:abl}
\end{figure*}

\textbf{Ablations on real-world peer-review environment.} To assess the effectiveness of reward from a simulated real-world peer-view environment, we ablate AlphaResearch-RM-7B at the first 400 iterations on \emph{``Packing Circles''} problem. 
\autoref{fig:abl} presents the execution results of w/ and w/o AlphaReasearch-RM-7B during the discovery process.
Compared to the baseline without RM, AlphaResearch-RM-7B successfully filtered 151 ideas below the threshold. This process yielded 108 correct rejections of execution failures while making 43 erroneous rejections of viable ideas.
AlphaResearch attained an accuracy of 71.5\% (108/151), a result that aligns closely with its  performance on the AlphaResearch-RM test set, as shown in \autoref{tab:rm-res}
This outcome effectively demonstrates the model's generalization capabilities and the efficacy of incorporating feedback from a simulated real-world peer-review environment.

\begin{figure}[t]
    \centering
    \includegraphics[width=0.85\linewidth]{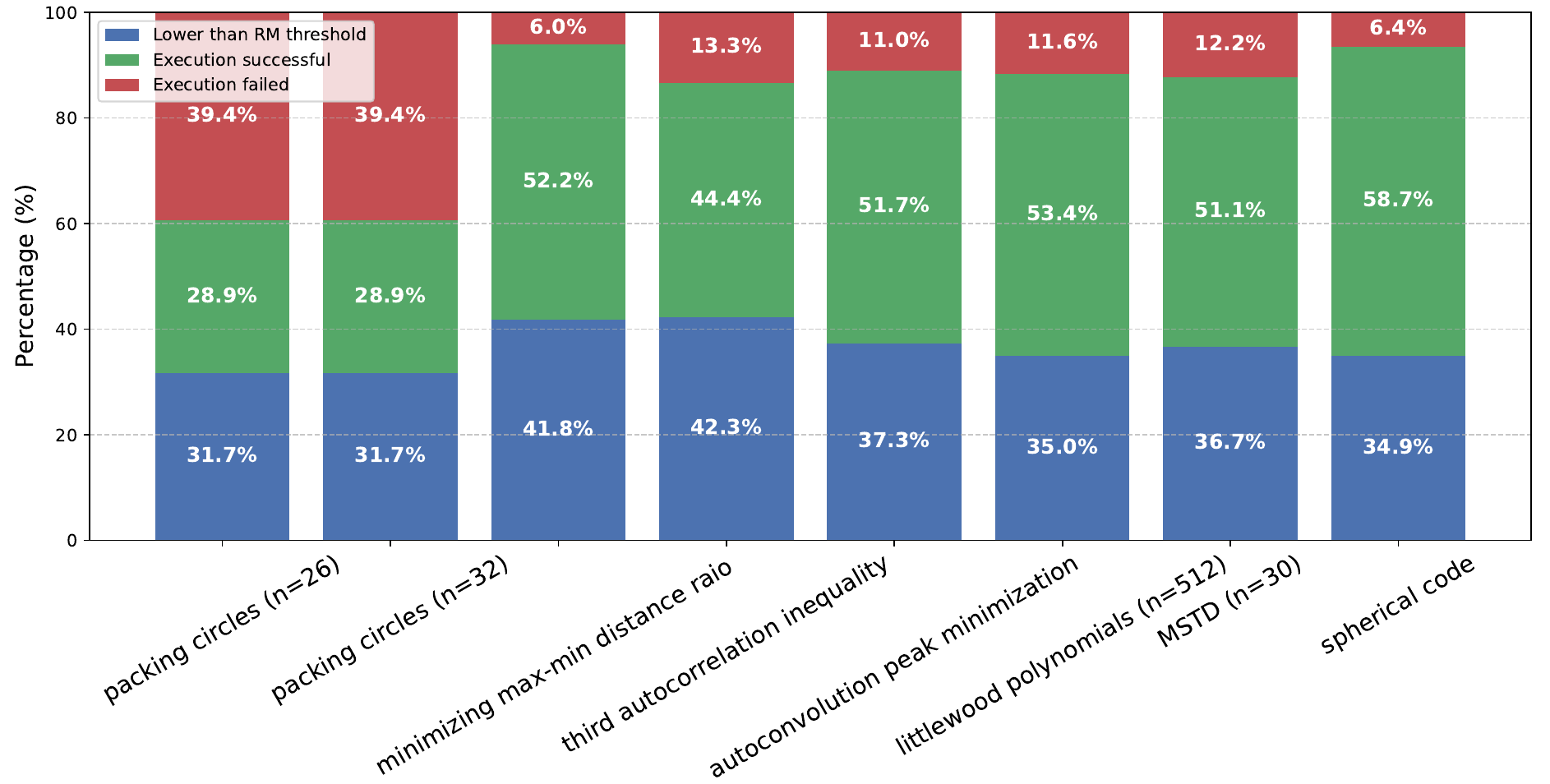}
    \caption{Reward overview during the discovery process. Each action in AlphaResearch will obtain 3 kinds of reward: (1) idea scrapping due to a lower RM score than the threshold (2) idea execution successes (3) idea execution fails.  }
    \vspace{-5pt}
    \label{fig:pr}
\end{figure}

\textbf{Analysis of the discovery process.} We analyze the rejection distribution in AlphaResearch discovery process. As shown in \autoref{fig:pr}, approximately 30\%$\sim$40\% of newly proposed ideas fall below the RM threshold and are thus discarded. The remaining ideas are executed, with the success rate of execution largely depending on the inherent characteristics of the problems. For example, the execution success rate on \emph{``Packing Circles''} problem is 28.9\%, whereas it reaches 51.7\% on the \emph{``Third Autocorrelation Inequality''} problem. \autoref{fig:reward} illustrates the execution-based rewards for these two examples in AlphaResearch. Despite the substantial variations in execution success rates, the execution-based rewards in both cases exhibit a consistent increasing trend. These findings demonstrate the interactions between LLM-based autonomous research agents and real-world environments.

\subsection{Case Study}
We select the successful example from AlphaResearch to better understand the discovery process. We’ll consider the problem \emph{``Packing Circles''} where the goal is to pack $n$ disjoint circles inside a unit square so as to maximize the sum of their radii, shown in \autoref{fig:example}. We first initialize AlphaResearch with an original research proposal and a related program that returns a list of circles $(x,y,r)$ as output. 
The verification program first employs \texttt{verify\_circles} function to check if the outputs of the initial program meet the problem constraints (e.g., all circles are inside a unit square) and \texttt{evaluate} function to output the sum of their radii.
The metadata, including: (1) research ideas, (2) programs, (3) execution results, are subsequently preserved as candidates which represent the end of one step.
At the next step, AlphaResearch will sample from the candidate pool and generate a new idea  to improve the research proposals from the sampled metadata.
After generating the new research ideas, AlphaResearch will further generate a patch to modify the existing program if the idea obtains a positive score from AlphaResearch-RM. The new program is then evaluated by the same verification program, thereby generating new metadata. We select the best program and idea as the final solution of AlphaResearch in this iterative process.

% \vspace{-0.3cm}
% \newpage

\section{Conclusion}
% \textbf{Conclusion.}
We present AlphaResearch, an autonomous research-oriented coding agent that synergistically combines new idea generation with program-based verification for autonomous algorithm discovery. 
To accelerate the discovery process of AlphaResearch, we construct a dual research-based environment to reduce ineffective attempts and runtime,. 
% Our approach also demonstrates that while scaling up attempts can facilitate algorithmic discovery, large-scale algorithm discovery still remains a challenge for current LLMs. 
On our collected 8 open-ended algorithmic problems, AlphaResearch outperforms AlphaEvolve for 2/8 algorithmic problems and demonstrates better discovery ability than previous state-of-the-art evolutionary agents, which demonstrates the effectiveness of interleaved idea generation and program generation and constructed peer-review environments in AlphaResearch. Furthermore, our systematic analysis of incorporating autonomous review feedback into autonomous discovery systems provides valuable insights for future research, contributing to the development of more advanced and versatile agentic discovery systems.

\bibliography{colm2026_conference}
\bibliographystyle{colm2026_conference}

\appendix
\addtocontents{toc}{\protect\setcounter{tocdepth}{3}}

% \renewcommand{\contentsname}{\Large Appendix Contents}
% \hypersetup{linkcolor=black}
% \tableofcontents
\newpage

\section{Limitations} 
While AlphaResearch demonstrates promising results in using a reward model (RM) to guide scientific idea search and generation, several important limitations should be acknowledged.
First, the AlphaResearch-RM is trained exclusively on ICLR peer-review records, which are heavily concentrated in machine learning and related topics. This creates a potential topic bias, where the model may perform less reliably when evaluating ideas outside core ML domains. Future work will explore training on more diverse and extensive peer-review datasets across broader scientific fields to mitigate this bias.
Second, there exists a fundamental mismatch between peer-review scores (which the RM is trained to predict) and the actual downstream usefulness or impact of a research idea. Peer review primarily assesses perceived novelty, technical soundness, and clarity at the time of submission, but does not necessarily reflect long-term scientific value, feasibility of execution, or practical utility. Therefore, optimizing for RM scores may not fully align with optimizing for genuinely useful or groundbreaking research.
Third, the current benchmark scope remains relatively narrow and partially inherited from existing datasets. Most evaluations focus on standard machine learning tasks and benchmarks, which may not fully capture the challenges of open-ended scientific discovery in more complex or interdisciplinary settings.
We believe addressing these limitations represents important directions for future research in building more reliable and general-purpose AI systems for scientific discovery.

\section{The Use of Large Language Models}
During the preparation of this manuscript, we utilized large language models (LLMs) for grammar checking and writing suggestions to enhance the readability and clarity of the content.

% \section{Experiment Cost}
% In this section, we present the experiment parameters (iterations,computational cost) required to reach the best solution for each on 8 tasks of AlphaResearchComp.
% \begin{table}[h]
% \centering
% \small
% \caption{Experiment Parameters of AlphaResearch .}
% \begin{tabular}{lcc}
% \toprule
% \textbf{Problem}                                         & \textbf{Iterations} &\textbf{ Cost per iteration (dollar)} \\
% \midrule
% packing circles (n=26)                          & 4768       & 0.013           \\
% packing circles (n=32)                          & 4768       & 0.013           \\
% minizing max-min distance ratio (d=2, n=16)      & 4400       & 0.017           \\
% third autocorrelation inequality                & 1366       & 0.012           \\
% autoconvolution peak minimization (upper bound) & 979        & 0.013           \\
% littlewood polynomials (n=512)                  & 2233       & 0.011           \\
% MSTD (n=30)                                     & 2826       & 0.011           \\
% spherical code                                  & 1132       & 0.015           \\
% \bottomrule
% \end{tabular}
% \end{table}

\section{Other Details}
\begin{figure}[h]

    \centering
    \includegraphics[width=0.49\linewidth]{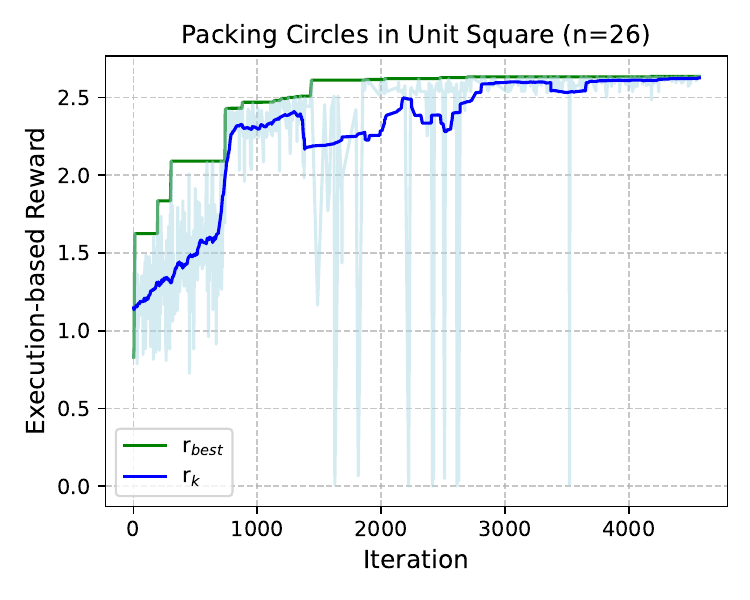}
    \includegraphics[width=0.49\linewidth]{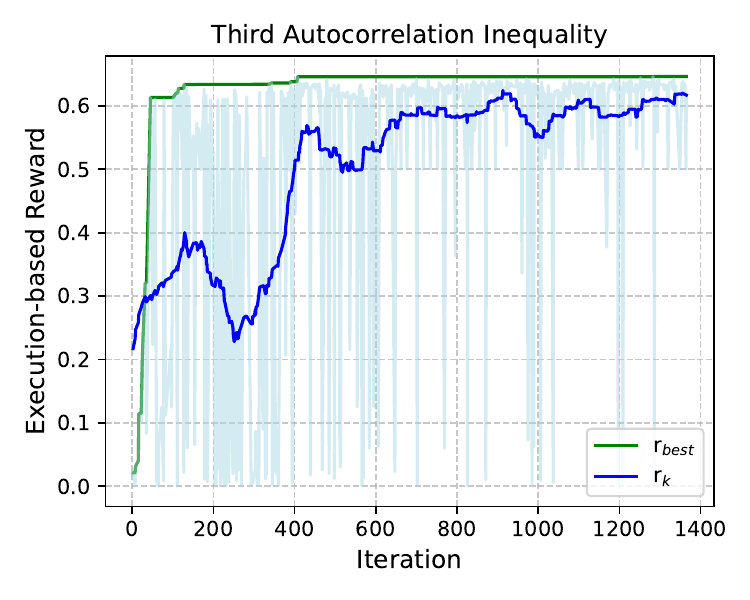}
    \caption{Execution-based reward of AlphaResearch on packing circles (n=26) problem (left) and third autocorrelation inequality problem (right).  }
    \label{fig:reward}
\end{figure}

\textbf{LLMs can refine their research ideas autonomously.} AlphaResearch discovers advanced algorithms by iteratively proposing and verifying new research ideas. As shown in \autoref{tab:rm-res}, 6/8 problems demonstrate consistent improvement throughout the discovery process. 
\autoref{fig:reward} presents two examples of the reward trend in AlphaResearch, where the execution-based reward initially grows rapidly, then slowly plateaus for optimal performance seeking.
This improvement trend emphasizes the autonomous discovery ability of research agents.

\begin{figure}[h]

    \centering
    \includegraphics[width=0.45\linewidth]{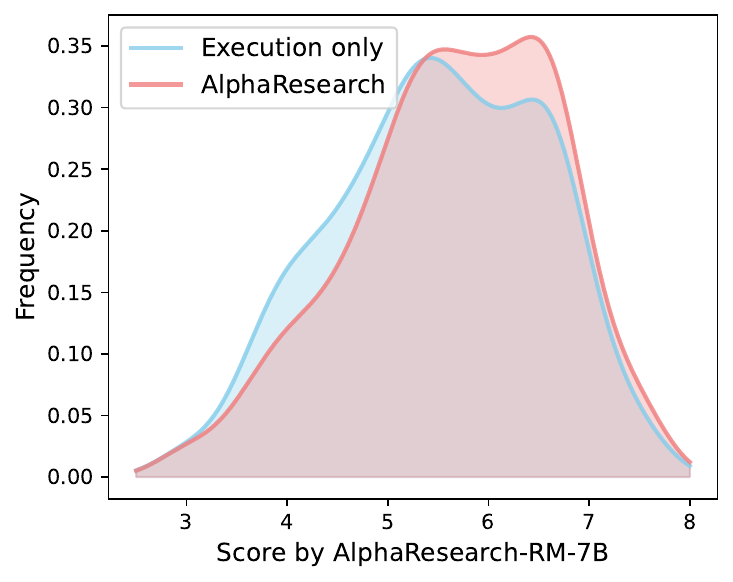}
    \caption{The idea comparison between the execution-only research agent and AlphaResearch, where AlphaResearch-RM-7B is used. This is done between the full distribution of all 1000 generated ideas from both agents without filtering.}
    \label{fig:dis}
\end{figure}

\paragraph{Execution-only agent against AlphaResearch.} To compare AlphaResearch with execution-only agents,  we utilize AlphaResearch-RM-7B to evaluate the novelty of ideas generated by the execution-only agent and ideas produced by AlphaResearch. As illustrated in \autoref{fig:dis}, the ideas generated by AlphaResearch generally achieve higher scores than execution-only research agents. This illustrates that AlphaResearch tends to generate better ideas to get higher external rewards, thus facilitating a more effective research optimization process.

% \section{Comparison with ShinkaEvolve}

% \textcolor{blue}{As shown in \autoref{fig:teaser}, we compare AlphaResearch, OpenEvolve \citep{openevolve} and ShinkaEvolve \citep{lange2025shinkaevolve} on \textit{packing circles (n=26)} problem at the first 500 steps for simplicity.
% AlphaResearch achieves better performance than OpenEvolve and slightly surpasses ShinkaEvolve, which demonstrates that dual research environments  could help research agent for scientific discovery.
% }

% ==================================================
% \newpage
% \section{Case Study during Discovery Process}
% \label{sec:ca}
% In the rejected pair from checkpoint 634, the revised draft 4f4c7847 is effectively identical to its parent e436c26a. Notably, this is found in the later period of the discovery process (Round 632-633). Aside from inflating Genetic Algorithm (GA) hyperparameters (e.g., population = 300 → 500, generations = 40 → 120) and adding an optional differential\_evolution branch, the entire pipeline above find\_better\_c3\_upper\_bound is byte-for-byte the same. Crucially, the core loop still calls the undefined normalize\_population, triggering the same NameError before any new logic can run. Because this “revision” neither fixes the blocking bug nor implements the promised multi-phase CMA-ES/surrogate/SOS pipeline, it constitutes only a cosmetic variant rather than a substantive new direction.

\begin{figure}[h]
    \centering
    \small
    \includegraphics[width=0.6\linewidth]{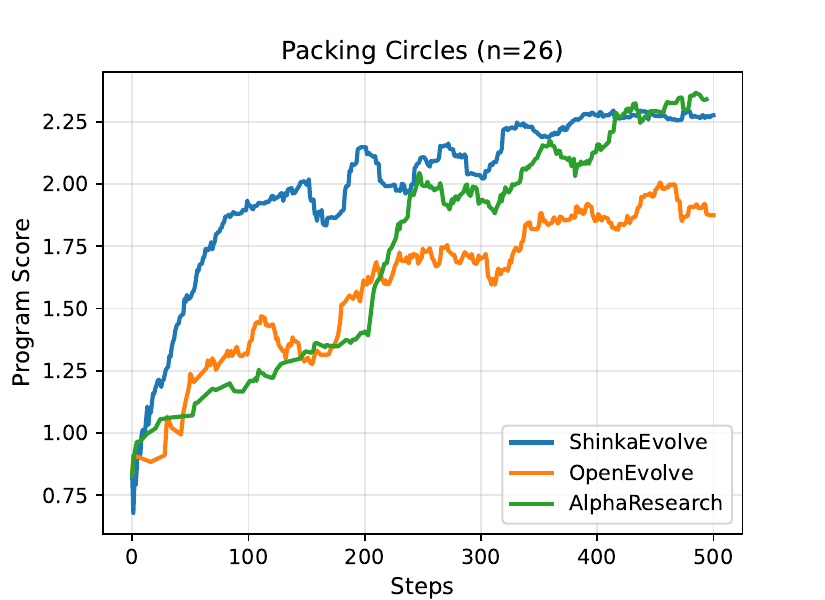}
    \caption{Comparison of OpenEvolve (with program-based reward), ShinkaEvolve (with program-based reward) and AlphaResearch (with program-based and peer-review reward). We run three agents on Packing Circles (n=26) problems. We compare AlphaResearch, OpenEvolve \citep{openevolve} with ShinkaEvolve \citep{lange2025shinkaevolve} on \textit{packing circles (n=26)} problem at the first 500 steps for simplicity.
AlphaResearch achieves better performance than OpenEvolve and slightly surpasses ShinkaEvolve, which demonstrates that dual research environments could help research agent for scientific discovery. }
    \label{fig:teaser}
\end{figure}

\section{Examples of Packing Circles}
\label{app:examples}
We show an example of the constructions discovered by AlphaResearch on problem \emph{``Packing Circles''}.

\begin{figure}[h]
    \centering
    \includegraphics[width=0.47\linewidth]{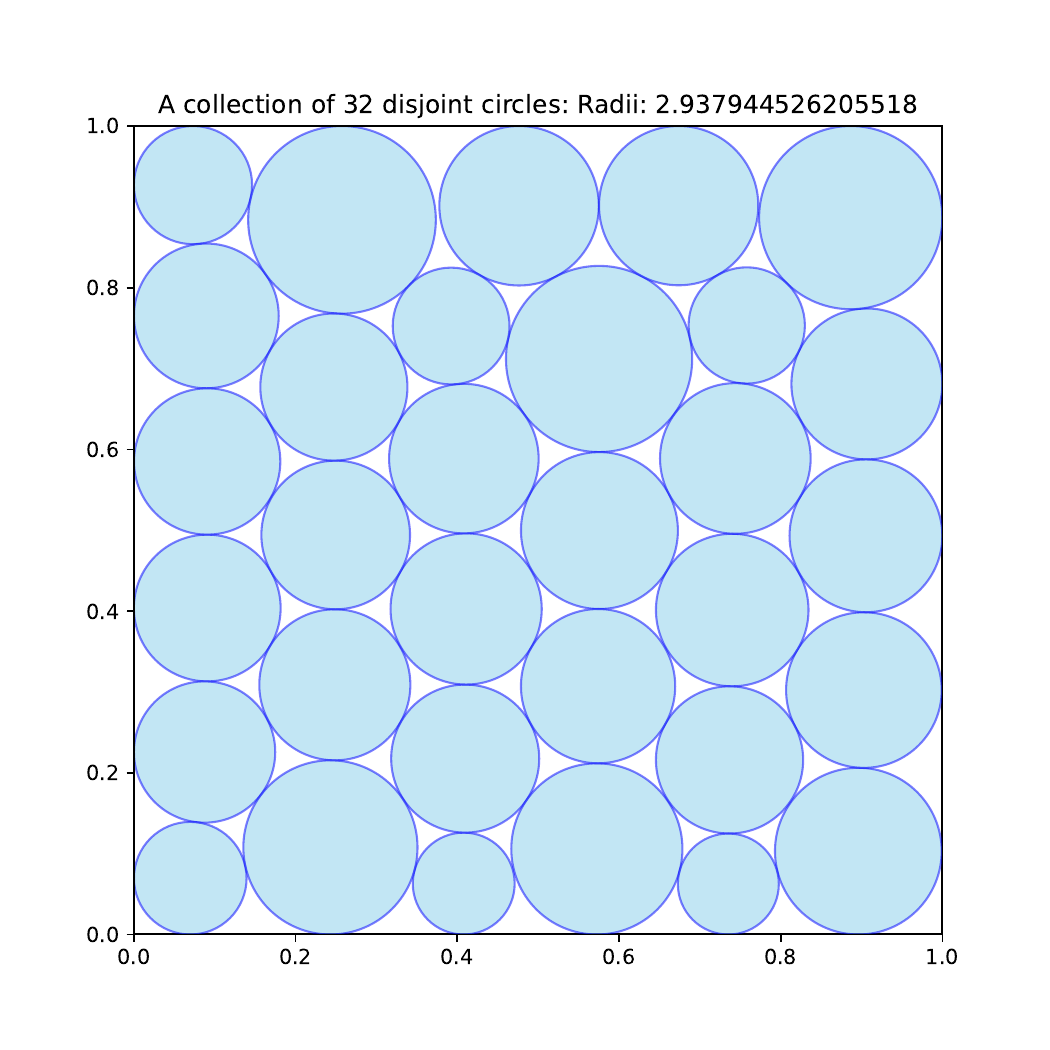}
    \includegraphics[width=0.47\linewidth]{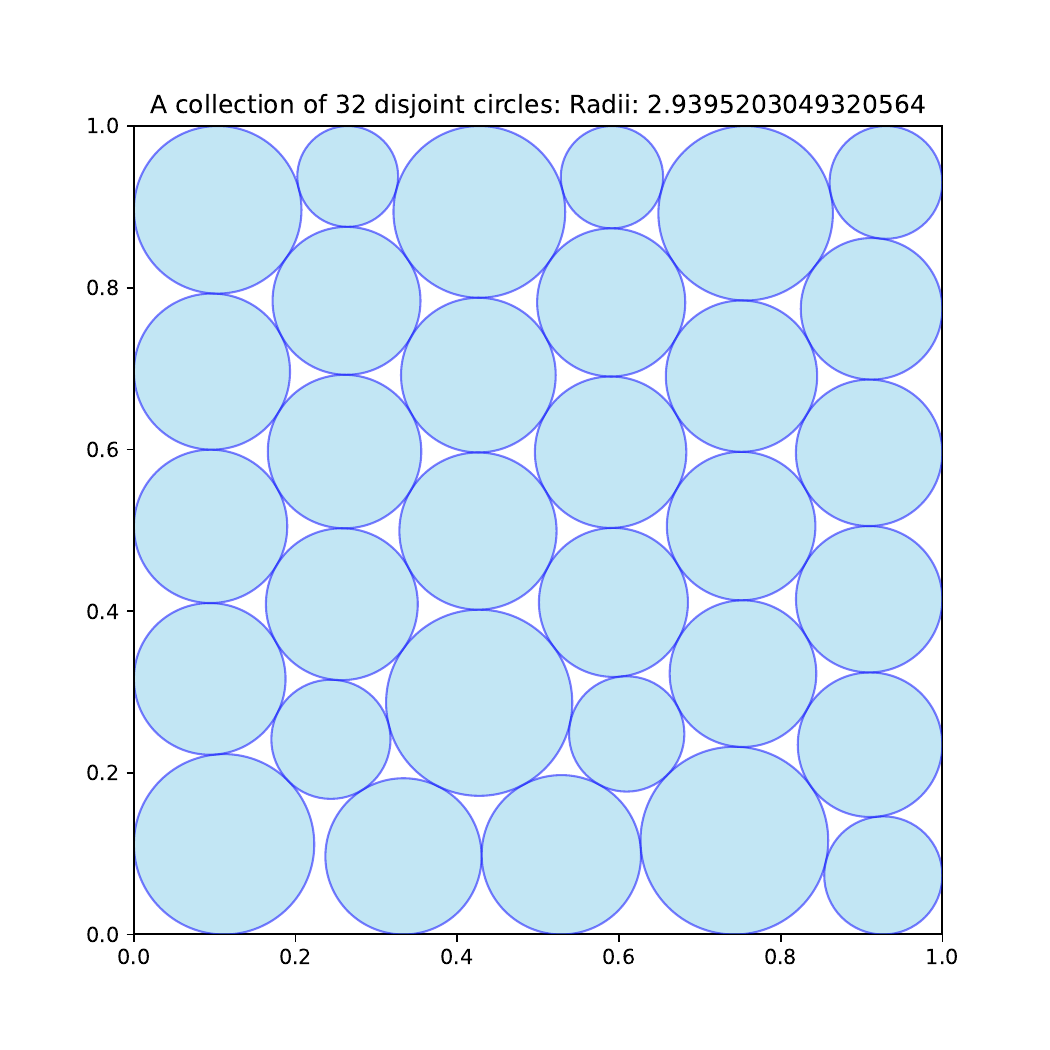}
    \caption{New construction of AlphaResearch (right) improving the best known AlphaEvolve (right) bounds on packing circles to maximize their sum of radii. Left: 32 circles in a unit square with sum of radii $\geq$ 2.9379. Right: 32 circles in a unit square with sum of radii $\geq$ 2.9395}
    \label{fig:comp-alpha}
\end{figure}

\begin{figure}[h]
    \centering
    \includegraphics[width=\linewidth]{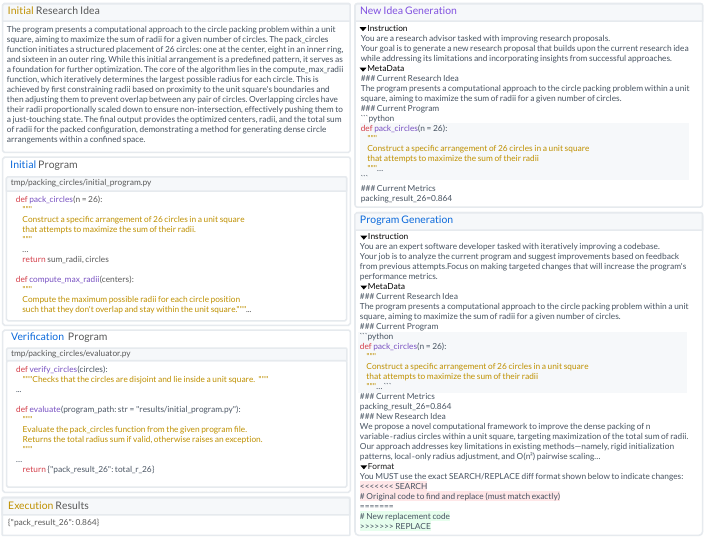}
    \caption{We show an example of a formatted task of AlphaResearch.}
    \label{fig:example}
\end{figure}

\end{document}